\documentclass[conference]{IEEEtran}
\IEEEoverridecommandlockouts
\usepackage{amsmath,amssymb,amsfonts}
\usepackage{algorithmic}
\usepackage{graphicx}
\usepackage{textcomp}
\usepackage{xcolor}
\usepackage{biblatex} %Imports biblatex package
\usepackage{booktabs}
\usepackage{multirow}

\def\BibTeX{{\rm B\kern-.05em{\sc i\kern-.025em b}\kern-.08em
    T\kern-.1667em\lower.7ex\hbox{E}\kern-.125emX}}
\begin{document}

\title{PRISM: A Transformer-based Language Model of Structured Clinical
Event Data\\

}

\author{\IEEEauthorblockN{Lionel Levine MS}
\IEEEauthorblockA{
\textit{UCLA}\\
Los Angeles, USA \\
orcidID 0000-0002-6926-7438}
\and
\IEEEauthorblockN{Dr. John Santerre}
\IEEEauthorblockA{\textit{UC Berkeley}\\
Berkeley, USA }
\and
\IEEEauthorblockN{Dr. Alex S. Young MD}
\IEEEauthorblockA{\textit{UCLA David Geffen School of Medicine}\\
Los Angeles, USA \\
orcidID 0000-0002-9367-9213}
\and
\IEEEauthorblockN{Dr. T. Barry Levine MD}
\IEEEauthorblockA{
\textit{ABLE Medical)}\\
Savannah, USA }
\and
\IEEEauthorblockN{Dr. Francis Campion MD}
\IEEEauthorblockA{
\textit{MITRE Corp.}\\
Bedford, USA \\
orcidID 0000-0002-0757-9305}
\and
\IEEEauthorblockN{Dr. Majid Sarrafzadeh}
\IEEEauthorblockA{
\textit{UCLA}\\
Los Angeles, USA}
}

\maketitle

\begin{abstract}
We introduce PRISM (Predictive Reasoning in Sequential Medicine), a transformer-based architecture designed to model the sequential progression of clinical decision-making processes. Unlike traditional approaches that rely on isolated diagnostic classification, PRISM frames clinical trajectories as tokenized sequences of events — including diagnostic tests, laboratory results, and diagnoses — and learns to predict the most probable next steps in the patient diagnostic journey. Leveraging a large custom clinical vocabulary and an autoregressive training objective, PRISM demonstrates the ability to capture complex dependencies across longitudinal patient timelines. Experimental results show substantial improvements over random baselines in next-token prediction tasks, with generated sequences reflecting realistic diagnostic pathways, laboratory result progressions, and clinician ordering behaviors. These findings highlight the feasibility of applying generative language modeling techniques to structured medical event data, enabling applications in clinical decision support, simulation, and education. PRISM establishes a foundation for future advancements in sequence-based healthcare modeling, bridging the gap between machine learning architectures and real-world diagnostic reasoning. 
\end{abstract}


\section{Introduction}
Accurate and timely clinical decision-making is fundamental to high-quality patient care. Traditionally, efforts to model diagnostic processes have relied on static classification frameworks, treating patient data as unordered feature sets aimed at predicting singular outcomes. However, real-world clinical reasoning unfolds dynamically: tests are ordered, results are evaluated, and diagnoses are formed through sequential, context-dependent decision-making processes.\cite{9380633}

To better capture this inherent structure, we propose PRISM (Predictive Reasoning in Sequential Medicine), a transformer-based framework that models clinical workups as tokenized sequences of events. Inspired by autoregressive language modeling, PRISM learns to predict the next likely diagnostic clinical action — whether a diagnostic test, laboratory result, or diagnosis — based on the evolving context of a patient's diagnostic timeline. By framing medical decision-making as a sequential generation problem, PRISM moves beyond outcome classification to simulate the dynamic reasoning patterns exhibited by clinicians.

Leveraging a customized clinical vocabulary and training on structured event data, PRISM demonstrates the ability to internalize complex patterns in diagnostic workflows. Early experimental results indicate that PRISM can accurately predict future clinical events across diverse patient trajectories, suggesting potential applications in clinical decision support, anomaly detection, workflow optimization, and medical education. This work establishes a foundational step toward generative modeling of healthcare processes, highlighting the power of modern transformer architectures to emulate aspects of clinical reasoning at scale.

\section{Background}
\subsection{Limits of Probabilistic Diagnostic Testing}

While Bayesian methods provide a mathematically rigorous framework for updating disease probabilities based on diagnostic tests, they (BLANK: struggle with higher order correlations between tests and the specific diagnoses.) 
 often rely on pairwise likelihoods—probabilities that relate specific tests individually to specific diagnoses. Although individually these probabilities can be empirically accurate and useful, real-world diagnostic situations frequently involve multiple overlapping comorbidities, complex interactions between test outcomes, and numerous patient-specific confounding variables and specific histories.

Early computerized diagnostic aids such as QMR‑DT relied on expert rules and Bayesian reasoning, but they covered fewer than 600 diseases and required years of manual curation.\cite{jaakkola1999variational}. Yet even painstakingly curated models like the QMR-DT, their performance degrades when the model is deployed in a new hospital whose disease prevalence and practice patterns differ from the original cohort.\cite{young2022empirical} Moreover, classical Bayesian networks are brittle at run time: they do not adapt gracefully when previously unseen evidence types or updated laboratory ranges are introduced.

More fundamentally, when multiple factors coexist, the assumption of conditional independence—often a critical prerequisite for practical Bayesian calculations—breaks down. The intricate interdependencies among symptoms, test results, and diseases in an individual patient’s unique clinical context make it computationally infeasible, and often conceptually inaccurate, to comprehensively enumerate and correctly weigh all conditional probabilities. Consequently, Bayesian approaches become limited in their ability to faithfully represent realistic uncertainty, since accurately calculating a truly reflective posterior probability distribution across all potential combinations of conditions and outcomes quickly becomes impossible.

Therefore, relying solely on Bayesian statistical modeling to drive diagnostic decision-making risks oversimplifying patient-level complexity. 

\subsection{Elaborating the Analogy Between Diagnostic Systems and LLM Token Prediction}

In language models (LLMs), each token—representing a word or sub-word unit—has clear pairwise probabilities relative to other tokens. For example, given a token such as "New," the subsequent token might have high probability for "York," "Zealand," or "Jersey." Individually, these pairwise probabilities are straightforward. However, as the sequence of tokens expands—considering an entire preceding sentence or paragraph—the combinatorial complexity and contextual dependencies make it impossible to precisely calculate probabilities using first order correlations alone. Hence, deep learning and attention mechanisms in transformer architectures (like GPT) approximate these complex conditional dependencies, capturing long-range contexts and intricate relationships between tokens.

By analogy, diagnostic medicine faces a similar complexity. Individual diagnostic tests may exhibit well-defined pairwise probabilities for specific diseases—just as pairwise token probabilities can be well-established. However, when attempting to predict the next best test or disease diagnosis given the entire preceding sequence (the patient's complete history, including medical tests, symptoms, prior diagnoses, lifestyle factors, and comorbidities), the scenario becomes profoundly more complex. The interactions between these elements are intricate, context-dependent, and frequently nonlinear, making exact probabilistic inference practically infeasible.

Thus, Bayesian statistical approaches quickly reach limitations. Transformer-based diagnostic models, analogous to language models, can approximate the complex contextual dependencies inherent in patient histories, making them potentially more capable of accurately predicting subsequent medical events (tests, results, or diagnoses). These models implicitly learn from data to reflect patient-level complexity more naturally, offering both predictive capability and interpretable attention scores to highlight influential factors in diagnostic reasoning.

\subsection{Prior Work in Transformer and Deep Learning Models for Sequential Clinical Diagnosis}

%%% ----------  Section 3 – Transformer-Based Approaches ----------

In contrast to high-level summations of pre-existing efforts, which are considerable and highly varied in nature, the authors here opt to focus in on prior work of immediate relevance to our proposed approach, by zeroing in on a methodology-specific approaches that have informed our own experimental design, and serve as a basis for which our work attempts to build on. 

\subsubsection{Tokenization and Representation of Medical Data}
\label{sec:tokenization}

The first step in adapting transformer models to diagnostic-sequence prediction is
to convert a patient’s richly structured history into a sequence of discrete
\emph{tokens}.  Early work relied on standardised vocabularies such as
ICD-9/ICD-10, CPT and Read codes, treating each code as an individual token
\cite{medalbert_2024,lotusai_2025}.  A coarser alternative aggregates thousands
of raw codes into a few hundred clinically coherent groups, reducing sequence
length while retaining clinical meaning \cite{medalbert_2024}.

More granular pipelines tokenise free-text concepts, symptoms and history
fragments extracted from clinical narratives \cite{autoddx_2024}.  The ETHOS framework extend this idea further by encompassing: admissions, diagnoses (ICD-10-CM), procedures (ICD-10-PCS),
medications (ATC), laboratory results (LOINC) and even inter-event time gaps. Under the ETHOS framework, these events are
mapped to 1–7 tokens each, then ordered chronologically into a patient-health
timeline (PHT) \cite{ethos_2024}.

Because multiple events can occur simultaneously during a visit, several authors
represent a patient record as a \emph{sequence of sets}, forcing the model to be
order-invariant within each set.  DPSS is the canonical example of this strategy
\cite{dpss_2020}.  Finally, semantic enrichment with external knowledge graphs
(or hierarchical ontologies such as CCS) can be injected by concatenating
concept embeddings with ontology embeddings, as done in the SETOR framework
\cite{setor_2023}.

\subsubsection{Transformer Model Architectures in Diagnostic Prediction}
\label{sec:architectures}

Traditionally model choice varied depending on the prediction task.  Encoder-only families (BERT,
ALBERT, MedAlbert) excel at classification problems such as early disease
detection from longitudinal code streams; MedAlbert, for instance, uses a
6-layer ALBERT encoder to flag incipient lung cancer three years in advance
\cite{medalbert_2024}.  Decoder-only designs (GPT-like) suit generative
forecasting: ETHOS, for instance, employs a causal decoder to roll out future PHTs in a zero-shot
setting \cite{ethos_2024}.  Hybrid encoder–decoder stacks (e.g.,
TransformEHR) translate past encounters into predicted future ICD sequences
\cite{lotusai_2025}.

Domain-aware modifications are common.  SETOR feeds visit-level embeddings—
augmented with CCS-derived ontology vectors and continuous-time positional
encodings—into a multi-layer transformer encoder dubbed the \emph{Patient
Journey Transformer} \cite{setor_2023}.  Such customisations can improve both
performance and interpretability without abandoning the core attention
machinery of the original architecture.

\subsubsection{Datasets and Evaluation Strategies}
\label{sec:datasets}

Most studies train on large, publicly available EHR corpora—chiefly
MIMIC-III/IV, which contain >200 k ICU and ward admissions with time-stamped
diagnoses, procedures, labs and medications.  Task-specific cohorts are also
common: WSIC North-West London primary-care records underpin MedAlbert’s lung
cancer work \cite{medalbert_2024}, while the DDXPlus collection provides
49-disease differential labels for multi-label diagnosis models
\cite{autoddx_2024}.  Limited access to longitudinal, high-quality datasets
remains a bottleneck \cite{setor_2023}.

Evaluation metrics mirror task formulations.  For single- or multi-label
classification, researchers report accuracy, precision, recall, F$_1$ and
AUROC; rank-based measures such as accuracy@$k$ gauge whether the true diagnosis
lies in the top-$k$ suggestions \cite{setor_2023}.  Generative timelines are
judged by downstream outcomes (e.g.\ mortality AUC, LOS MAE) or exact-match
token accuracy \cite{ethos_2024}.  Standard practice splits data into
train/validation/test partitions, sometimes augmented with cross-validation or
zero-shot evaluations on unseen institutions to probe generalisability
\cite{dpss_2020}.

\section{Methods}
For this framework, we employed data from the publicly available MIMIC-IV database, an extensive electronic health records repository collected from patients admitted to the Beth Israel Deaconess Medical Center (BIDMC) between 2008 and 2019. The database encompasses detailed clinical information, including demographics, vital signs, laboratory test results, diagnostic procedures, and discharge diagnoses captured during hospital stays \cite{johnson2023mimiciv}.

\subsection*{Patient Selection and Diagnostic Trajectory Filtering}

To investigate diagnostic trajectories, we chose a specific clinical use case to train the model on. Specifically, we opted for patients with initial presentations of undiagnosed chest pain followed by confirmed cardiac conditions. To obtain this subset, we constructed a filtering pipeline using structured data from the MIMIC-IV database. We selected patients whose first hospital encounter included a diagnosis of unspecified or non-specific chest pain, and who were subsequently diagnosed with a more definitive cardiac condition during their clinical course.

Patients were first filtered based on their earliest recorded ICD-9-CM diagnosis, specifically for unspecified or atypical chest pain. The following 5-digit ICD-9 codes were used to define initial inclusion:

\begin{itemize}
  \item \textbf{Initial Admission for Unspecified Chest Pain}
  \begin{itemize}
    \item \texttt{786.50} -- Chest pain, unspecified
    \item \texttt{786.51} -- Precordial pain
    \item \texttt{786.52} -- Painful respiration
    \item \texttt{786.59} -- Other chest pain
  \end{itemize}
\end{itemize}

Following this, patients were retained in the analysis if they subsequently received any diagnosis corresponding to a defined cardiac condition. The cardiac categories and their associated ICD-9-CM codes are listed in Figure 1.

\begin{figure}
    \centering
    \includegraphics[width=1\linewidth]{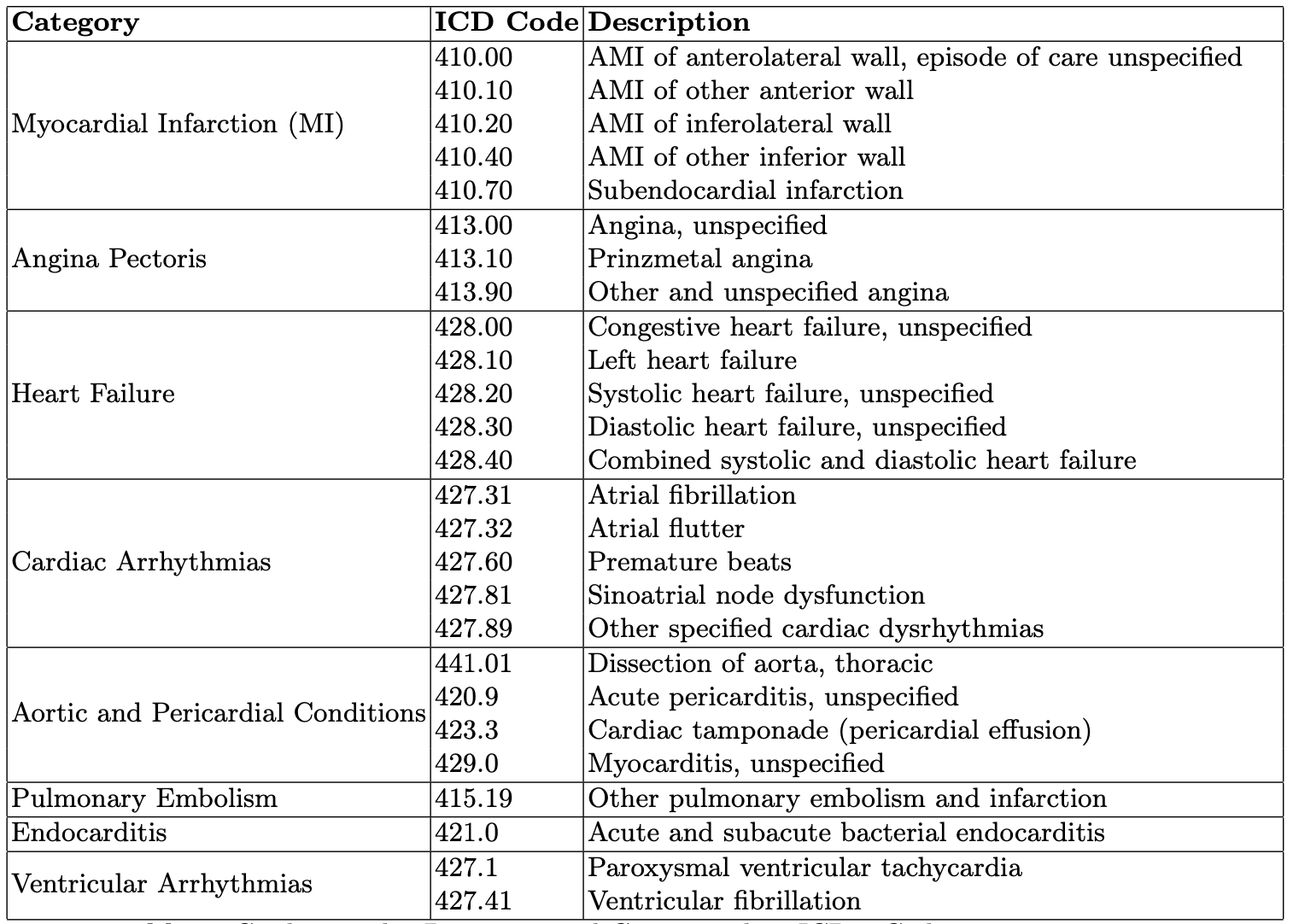}
    \caption{CARDIOVASCULAR DIAGNOSES AND CORRESPONDING ICD-9 CODES}
    \label{fig:icd-codes}
\end{figure}

A total of 14,536 unique patients were hospitalized with one of these initial diagnoses, and 3,164 patients were subsequently diagnosed with one of these 26 terminal diagnoses. Temporal ordering of events was verified using timestamped diagnosis records to ensure that cardiac diagnoses occurred after the initial chest pain admission. This filtered cohort was used for downstream analysis involving diagnostic test patterns, time-to-diagnosis evaluation, and treatment initiation windows.

\subsection*{Batch Tokenization of Patient Clinical Timelines}

To prepare structured clinical data for transformer-based modeling, we developed a batch tokenization pipeline that converts patient event histories into sequential token representations. This process standardizes heterogeneous clinical events into discrete, interpretable tokens suitable for autoregressive learning frameworks.

We extracted the following six event classes corresponding to tables in the MIMIC‑IV database:
\begin{enumerate}
    \item Demographics 
    \item Admissions/Discharges 
    \item Laboratory results 
    \item Outpatient measurements 
    \item Microbiology cultures 
    \item ICD‑coded diagnoses
\end{enumerate}

Each table was filtered for the target subject-ID, relabeled with clinically meaningful descriptors (e.g., itemid → test name), and assigned an event tag indicating its provenance.
Events were then vertically concatenated and sorted first by time and second by a fixed event precedence (admission → OMR → lab → microbiology → diagnosis → discharge), and subsequently, alphabetically within each event category to create a deterministic, chronological timeline.

\paragraph{Input Data.}
The pipeline accepts individual patient histories stored as CSV files, where each row represents a timestamped clinical event. Events span multiple categories, including demographic information, laboratory tests, diagnoses, medical observations, and microbiology reports. Each event type is associated with specific fields (e.g., lab test priority, diagnosis codes, observation results).

For this initial project, no therapeutic, pharmacological, or other clinical interventions were included, as the focus was narrowly on the process of disease identification and ultimate diagnosis. 

\paragraph{Tokenization Strategy.}
For each event, a corresponding token is generated according to predefined templates demonstrated in Figure 2.

\begin{figure}
    \centering
    \includegraphics[width=1\linewidth]{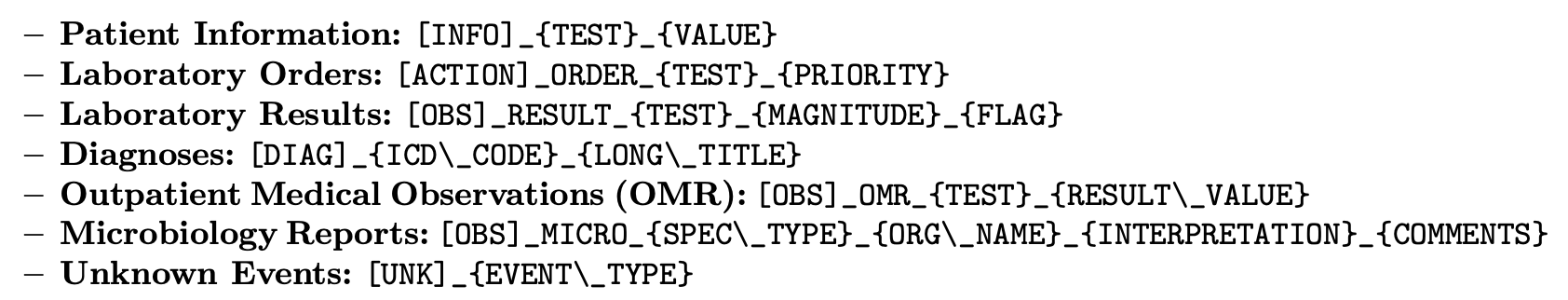}
    \caption{Tokenization Strategy for Clinical Events}
    \label{tokenization}
\end{figure}

All text fields are converted to uppercase, with spaces replaced by underscores for consistency. Missing or null values are imputed using placeholder tokens such as \verb|UNKNOWN|, \verb|NONE|, or \verb|NO\_COMMENTS|, depending on context.

\paragraph{Batch Processing.}
The pipeline recursively processes all CSV files within a specified input directory. For each patient file, a corresponding tokenized document is generated, where tokens are space-delimited and saved with a \verb|.txt| extension in a designated output directory.

\paragraph{Output.}
Each tokenized file captures the full chronological sequence of clinical events for a patient in token form, enabling downstream tasks such as next-action prediction, diagnostic trajectory modeling, and clinician behavior simulation. 

Notably, events with simultaneous timestamps, for instance a battery of tests ordered, or multiple concurrent diagnoses being input into and EHR, likely as a result of a batched input, were assigned an arbitrary ordering that results in the appearance of a sequential ordinality not necessarily reflective in the actual diagnostic progression. This limitation will be noted and explored further in the discussion section.

\subsection*{Vocabulary Construction for Clinical Token Sequences}

To facilitate transformer-based modeling of structured clinical data, we implemented a vocabulary construction pipeline that maps discrete clinical tokens to unique integer identifiers. This process ensures standardized input representation for autoregressive language models while controlling vocabulary size to optimize computational efficiency.

\paragraph{Token Collection.}
The vocabulary builder processes all tokenized patient documents within a specified directory. Each document contains space-delimited tokens representing chronological clinical events, including diagnostic tests, observations, diagnoses, and treatments.

\paragraph{Frequency-Based Pruning.}
To address the long-tail distribution inherent in clinical event data, we applied a frequency-based pruning strategy. Token frequencies were computed across the entire corpus, and the vocabulary was limited to the \textit{N} most frequent tokens, where \textit{N} denotes the predefined maximum vocabulary size (10,000 tokens in this case). This approach retains common clinical patterns while mapping infrequent or rare tokens to a universal unknown token.

\paragraph{Special Tokens.}
Two reserved tokens were incorporated:
\begin{itemize}
    \item \verb|[PAD]| : Represents padding for sequence alignment in batch processing.
    \item \verb|[UNK]| : Captures all tokens excluded due to frequency pruning or unseen during inference.
\end{itemize}
These tokens were assigned fixed indices (0 for \verb|[PAD]| and 1 for \verb|[UNK]|).

\paragraph{Vocabulary Output.}
The finalized vocabulary was serialized in JSON format, providing a mapping from token strings to integer indices. This dictionary serves as the reference for encoding tokenized clinical sequences into numerical tensors compatible with transformer architectures.

\subsection*{Model Architecture}

We employed a decoder-only transformer architecture inspired by GPT-2 to model sequential clinical events in an autoregressive framework. The model was configured to process tokenized representations of patient timelines, where each token corresponds to a discrete clinical action, observation, or diagnosis.

The transformer configuration included:
\begin{itemize}
    \item \textbf{Vocabulary Size:} 10,000 tokens (including special tokens \verb|[PAD]| and \verb|[UNK]|)
    \item \textbf{Maximum Sequence Length:} 512 tokens
    \item \textbf{Embedding Dimension:} 256
    \item \textbf{Number of Layers:} 6 transformer decoder blocks
    \item \textbf{Number of Attention Heads:} 8
    \item \textbf{Positional Encoding:} Learned embeddings
\end{itemize}

The model was instantiated using the HuggingFace Transformers library, with random weight initialization to accommodate the custom clinical vocabulary.

\subsection*{Training Procedure}

The model was trained using a causal language modeling objective, where the task is to predict the next token in a sequence given all preceding tokens. Cross-entropy loss was computed across all token positions, ignoring padded tokens.

\begin{itemize}
    \item \textbf{Optimizer:} AdamW with a learning rate of $5 \times 10^{-4}$
    \item \textbf{Batch Size:} 8 sequences
    \item \textbf{Number of Epochs:} 5
    \item \textbf{Loss Function:} CrossEntropyLoss (applied to next-token prediction)
    \item \textbf{Hardware:} Training conducted on a Python 3 Google Compute Engine backend (GPU) with 83.5 GB System RAM, 40 GB GPU RAM and 235.6 GB Disk space.
\end{itemize}

The dataset of tokenized patient timelines was split into training, validation, and test sets using an 80/10/10 ratio. Padding or truncation was applied to standardize sequence lengths to 512 tokens.

\subsection*{Validation and Checkpointing}

After each epoch, validation loss was computed across the held-out validation set. The model parameters were saved whenever a new lowest validation loss was achieved. This early stopping strategy ensured retention of the best-performing model without overfitting.

% ---------- 4. Results ----------
\section{Results}

\subsection{Training dynamics}

\begin{figure}[h]
  \centering
  \includegraphics[width=.75\linewidth]{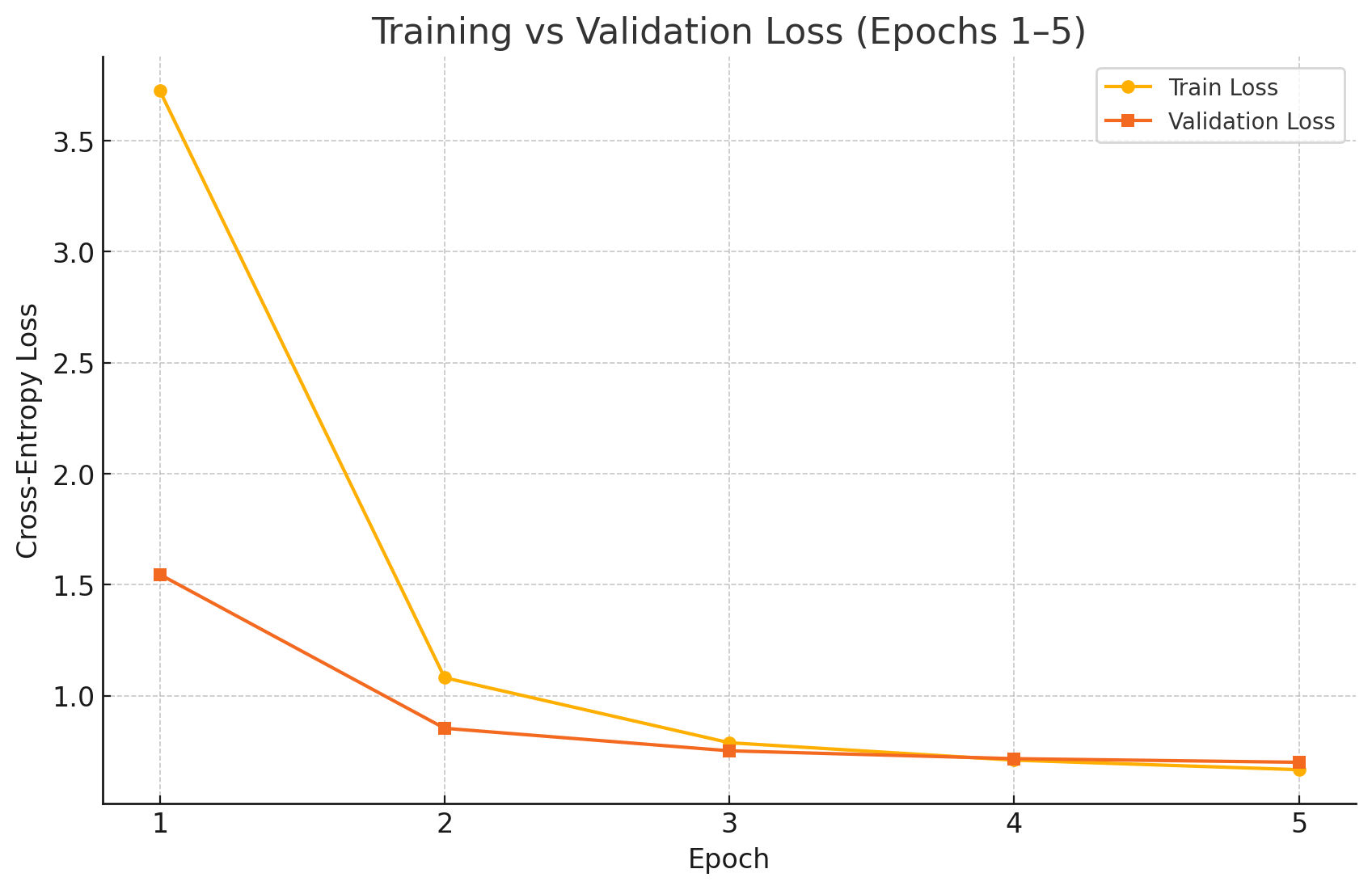}
  \caption{Cross‑entropy loss on training and validation data over five epochs.}
  \label{fig:learning_curves}
\end{figure}

Figure~\ref{fig:learning_curves} shows the cross‑entropy trajectories for the
decoder‑only transformer over five epochs.  Loss on both the training and
validation splits fell sharply during the first two epochs and then converged
to a shallow plateau, indicating rapid optimisation followed by stable
generalisation.  The minimal gap ($<\!0.03$) between training and validation
loss after epoch 3 suggests that regularisation noise (e.g.\ dropout and weight
decay) rather than over‑fitting accounts for the remaining discrepancy.

\subsection{Final model performance}

Table~\ref{tab:epoch_metrics} summarises the quantitative metrics.  The best
checkpoint, obtained at epoch 5, achieved a validation cross‑entropy of
\textbf{0.7000}, corresponding to a perplexity of \textbf{2.01}.  In other
words, conditioned on the patient’s history, the model narrows the
10\,000‑token clinical vocabulary to roughly two plausible next events on
average—removing more than 92 \% of the intrinsic sequence entropy.

\begin{table}[h]
  \centering
  \caption{Training and validation metrics per epoch.  Perplexity (PPL) is
    computed as $e^{\text{loss}}$.}
  \label{tab:epoch_metrics}
  \begin{tabular}{@{}cccccc@{}}
    \toprule
    \multirow{2}{*}{Epoch} & \multicolumn{2}{c}{Training} &
    \multicolumn{2}{c}{Validation} & \multirow{2}{*}{Best ckpt?} \\
    \cmidrule(lr){2-3}\cmidrule(l){4-5}
     & Loss & PPL & Loss & PPL & \\
    \midrule
    1 & 3.7257 & 41.5 & 1.5462 & 4.70 & -- \\
    2 & 1.0811 & 2.95 & 0.8532 & 2.35 & -- \\
    3 & 0.7881 & 2.20 & 0.7517 & 2.12 & -- \\
    4 & 0.7096 & 2.03 & 0.7163 & 2.05 & -- \\
    5 & 0.6666 & 1.95 & \textbf{0.7000} & \textbf{2.01} & X \\
    \bottomrule
  \end{tabular}
\end{table}

\subsection{Interpretation}

A validation perplexity of $\sim\!2.0$ means the model eliminates $\approx\!
11.3$\,bits of uncertainty per token compared with a uniform 10\,000‑way guess
($\log_{2}10\,000=13.3$), retaining only $\sim\!1$\,bit.  Such predictive
confidence is sufficient for practical decision‑support use‑cases (e.g.\
top‑$k$ next‑event suggestions) and provides dense clinical embeddings for
down‑stream risk‑stratification tasks.

While no explicit baseline models (e.g., n-gram predictors or recurrent neural networks) were implemented in this study, it is anticipated that such approaches would struggle to capture the long-range dependencies and complex branching logic present in clinical timelines. The transformer’s ability to handle extended context windows and model nuanced sequential relationships underscores its suitability for this domain.

These quantitative findings provide a assertive validation of the proposed tokenization strategy and model architecture, demonstrating that large-vocabulary, structured medical data can be effectively modeled using language modeling techniques. Beyond predictive performance, this foundation enables practical applications in clinical decision support, simulation of diagnostic processes, and identification of deviations from standard care pathways.

\subsection*{Next-Token Generation Experiments}

To evaluate the practical applicability of the trained transformer model, we
conducted a series of autoregressive next-token generation experiments. The goal
was to simulate clinician behavior by predicting subsequent diagnostic actions,
test outcomes, or diagnoses based on partial patient timelines.

Three distinct clinical scenarios were designed to assess the model's ability to:
\begin{enumerate}
    \item Predict likely diagnostic conclusions.
    \item Anticipate laboratory test outcomes.
    \item Suggest subsequent diagnostic tests in ongoing evaluations.
\end{enumerate}

For each scenario, an initial sequence of tokens reflecting patient demographics and early clinical actions was provided as a prompt. The model then generated a continuation of up to 15 tokens. The results are demonstrated in Figure 4.

\begin{figure}
    \centering
    \includegraphics[width=1\linewidth]{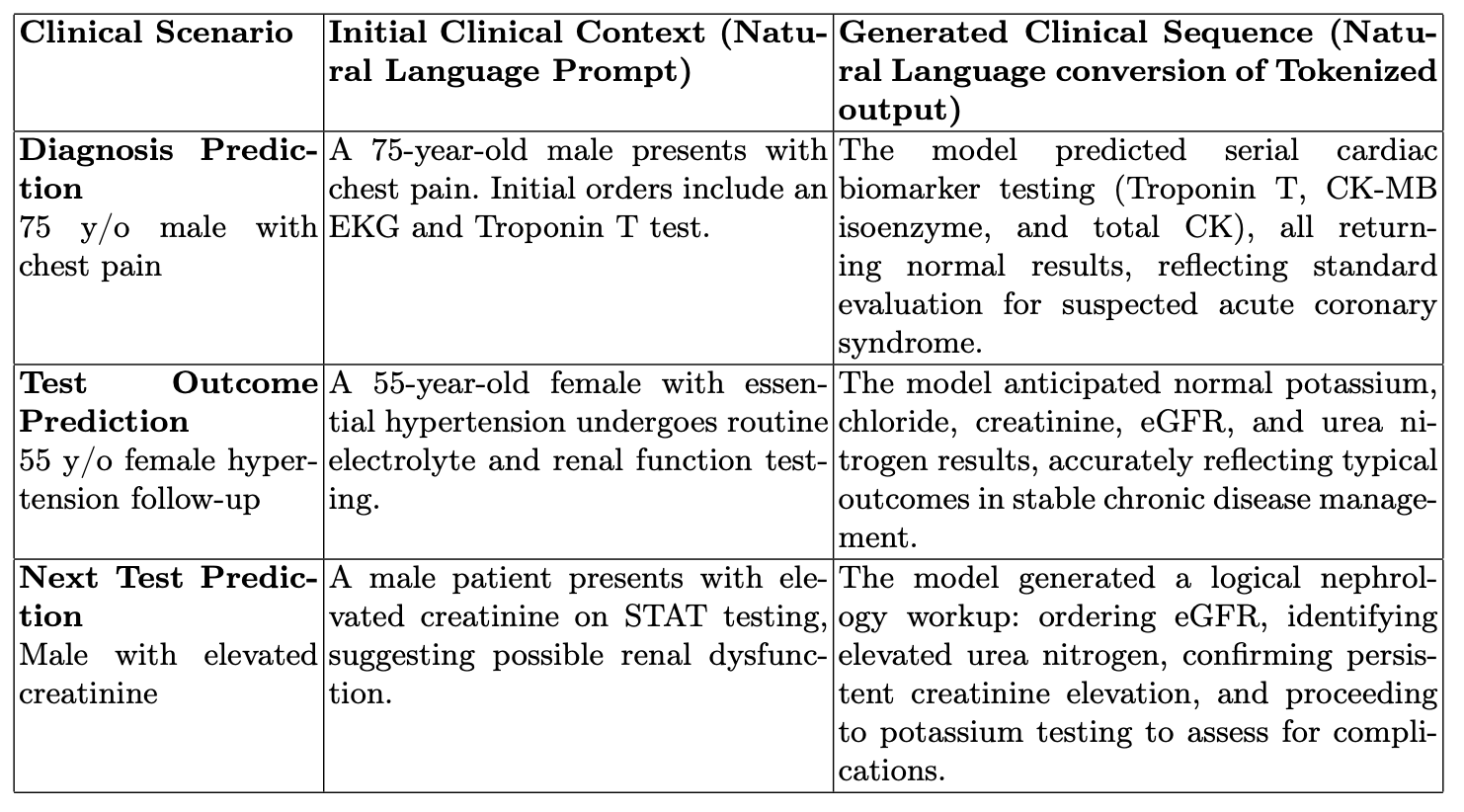}
    \caption{Examples of Next-Token Generation Across Clinical Scenarios}
    \label{fig:Clinical-scenarios}
\end{figure}

\subsection*{Interpretation of Generated Sequences}

The generated sequences demonstrated clinically coherent patterns aligned with standard diagnostic workflows:

\begin{itemize}
    \item In the \textbf{diagnosis prediction} scenario, the model appropriately simulated serial cardiac biomarker testing, reflecting common practice in evaluating suspected acute coronary syndrome.
    \item For \textbf{test outcome prediction}, the model correctly anticipated normal electrolyte and renal function results in a stable hypertensive patient.
    \item In the \textbf{next diagnostic test} scenario, the model followed a logical nephrology workup sequence after detecting elevated creatinine, including eGFR, BUN, and potassium testing.
\end{itemize}

These results highlight the model's capacity to internalize complex clinical reasoning patterns, suggesting potential utility in decision-support applications. Minor artifacts, such as repetitive test ordering and occasional inconsistencies in lab outcomes, were observed and are discussed in detail in the Discussion section.

\section*{Discussion}

The results of this study demonstrate that a transformer-based language model can effectively learn and predict sequential patterns within structured clinical event data. The PRISM model exhibited strong predictive performance not only for diagnostic conclusions but also for the sequencing of diagnostic tests and expected test outcomes. These findings support the viability of modeling clinical reasoning processes as an autoregressive token prediction task, uncovering latent structure in diagnostic workflows.

A notable strength of this approach lies in its ability to anticipate plausible clinical actions, even when faced with incomplete information. The model’s successful simulation of common diagnostic strategies, such as serial cardiac biomarker testing or stepwise nephrology workups, underscores its capacity to internalize complex clinical reasoning patterns. This has immediate implications for clinical decision support, where predictive modeling can guide test ordering, suggest next diagnostic steps, or identify deviations from expected care pathways.

\subsection{Limitations}
\subsubsection{1. Narrow Patient Cohort Scope}
To enhance signal fidelity and reduce the complexity of modeling heterogeneous patient trajectories, we deliberately constrained our training cohort. While over 14,000 patients initially presented with unspecified chest pain, only 3,164 with confirmed cardiac conditions were retained for modeling. This targeted subset dramatically narrowed the range of diagnostic trajectories, yielding cleaner data for learning but limiting the model’s generalizability to broader or more ambiguous clinical populations. As a result, PRISM will likely not yet generalize well to patients with non-cardiac pathologies or atypical presentations.

\subsubsection{2. Exclusion of Procedures and Medications}
For both conceptual clarity and computational tractability, we excluded therapeutic interventions such as procedures and medication administrations from the tokenized input sequences. This decision was motivated by the need to constrain the vocabulary to a manageable 10,000-token cap, minimizing the number of out-of-vocabulary events and improving model convergence. However, this exclusion restricts the model’s diagnostic reasoning capabilities—especially in clinical contexts where treatment response, procedural complications, or drug interactions significantly influence diagnostic trajectories. Consequently, PRISM may overlook important causal pathways or fail to contextualize diagnostic events relative to ongoing therapies.

\subsubsection{3. Linear Modeling of Co-Occurring Events}
The current model architecture assumes a strictly sequential ordering of all clinical events, which does not always reflect real-world practice. In many cases, clinicians order multiple tests simultaneously or enter multiple diagnoses in a batch. To accommodate this, events sharing a timestamp were assigned a fixed deterministic order based on event type and alphabetical sorting. While this approach enabled compatibility with autoregressive modeling, it introduces artificial ordering that may distort the temporal relationships between co-occurring events. This is particularly problematic in fast-paced or protocol-driven scenarios where decision-making occurs in parallel. The inability to represent sets of simultaneous or unordered events limits the fidelity with which PRISM can model genuine clinical workflows.

Future iterations of this work may address these limitations by broadening cohort inclusion criteria to encompass more diverse patient populations, integrating therapeutics into the event vocabulary using hierarchical compression or clinically abstracted embeddings, and adopting hybrid sequence-set or graph-based representations to better capture the multidimensional nature of clinical care. These modifications will further enhance PRISM’s ability to emulate clinician reasoning and support real-time diagnostic assistance in complex medical environments.

These findings have several implications. First, they highlight the potential of language models to approximate aspects of clinician behavior, offering avenues for decision-support systems that can anticipate subsequent diagnostic steps or flag deviations from standard workflows. Second, the ability to predict likely test results prior to their execution could inform opportunistic diagnostic strategies, optimizing resource utilization and reducing unnecessary testing.

\section*{Conclusion}

This study demonstrates the feasibility and effectiveness of applying transformer-based language modeling techniques to structured clinical event data. By leveraging a custom tokenization strategy and training a decoder-only transformer architecture on patient diagnostic timelines, we successfully captured complex sequential patterns inherent in clinical decision-making processes.

The model’s ability to achieve a substantial reduction in validation loss, far outperforming random baselines within a large vocabulary space, highlights the predictability embedded within standardized diagnostic workflows, laboratory testing sequences, and common clinical pathways. Furthermore, next-token generation experiments illustrated that the model could simulate plausible clinician behavior, anticipating diagnostic tests, test outcomes, and potential diagnoses in alignment with established medical practice.

These findings suggest that language models, when properly adapted, offer a powerful framework for modeling healthcare processes beyond traditional natural language tasks. Potential applications include clinical decision support, automated simulation of diagnostic reasoning, medical education tools, and identification of deviations from standard care pathways.

Future work will focus on enhancing the model’s handling of parallel clinical actions, reducing repetitive token generation through advanced sampling strategies, and integrating temporal context more explicitly. Additionally, incorporating clinician-in-the-loop evaluations and benchmarking against alternative modeling approaches will further validate the utility and robustness of this framework.

In summary, this research provides a foundational step towards leveraging generative transformer models to better understand and support clinical workflows, bridging the gap between machine learning capabilities and real-world healthcare decision-making.

\section{Credits}
This study was independently funded
by participating researchers. Special acknowledgment to the MIMIC project for making the MIMIC-IV database available for this research. 

\section{Disclosures}
The authors have no competing interests to declare that are
relevant to the content of this article. 

\printbibliography %Prints bibliography

@article{johnson2023mimiciv,
  author    = {Johnson, Alistair E. W. and Bulgarelli, Lucas and Shen, Lu and Gayles, Andre and Shammout, Ayah and Horng, Steven and Pollard, Tom J. and Moody, Benjamin and Gow, Benjamin and Lehman, Li-wei H. and Celi, Leo Anthony and Mark, Roger G.},
  title     = {{MIMIC-IV}, a freely accessible electronic health record dataset},
  journal   = {Scientific Data},
  volume    = {10},
  number    = {1},
  pages     = {1--18},
  year      = {2023},
  publisher = {Nature Publishing Group},
  doi       = {10.1038/s41597-022-01899-x},
  url       = {https://www.nature.com/articles/s41597-022-01899-x}
}

@ARTICLE{9380633,
  author={Mansour, Romany Fouad and Amraoui, Adnen El and Nouaouri, Issam and Díaz, Vicente García and Gupta, Deepak and Kumar, Sachin},
  journal={IEEE Access}, 
  title={Artificial Intelligence and Internet of Things Enabled Disease Diagnosis Model for Smart Healthcare Systems}, 
  year={2021},
  volume={9},
  number={},
  pages={45137-45146},
  doi={10.1109/ACCESS.2021.3066365}}

@article{autoddx_2024,
  title        = {Automatic Differential Diagnosis Using Transformer‐Based Multi-Label Sequence Classification},
  author       = {Anonymous},
  journal      = {arXiv preprint},
  year         = {2024},
  eprint       = {2408.15827},
  archivePrefix= {arXiv}
}

@article{ethos_2024,
  title        = {A Transformer-Based Model for Zero-Shot Health Trajectory Prediction},
  author       = {Anonymous},
  journal      = {medRxiv},
  year         = {2024},
  eprint       = {2024.02.29.24303512},
  archivePrefix= {medRxiv}
}

@article{medalbert_2024,
  title        = {Transformer-Based Deep Learning Model for the Diagnosis of Lung Cancer in Primary Care},
  author       = {Anonymous},
  journal      = {npj Digital Medicine},
  year         = {2024},
  note         = {Accessed 11 Apr 2025}
}

@misc{lotusai_2025,
  title        = {Transforming Disease Prediction with {LotusAI-Predict}: A Fine-Tuned {LLaMA} Model},
  author       = {Lotus Health AI},
  howpublished = {\url{https://lotushealth.ai/blog/transforming-disease-prediction-with-lotusai-predict}},
  year         = {2025},
  note         = {Accessed 11 Apr 2025}
}

@inproceedings{dpss_2020,
  title        = {Diagnostic Prediction with Sequence-of-Sets Representation Learning for Clinical Events},
  author       = {Zhang, Yue and Chen, Liwei and Ghosh, Debashis},
  booktitle    = {Proceedings of the ACM Conference on Health, Inference, and Learning},
  year         = {2020}
}

@article{setor_2023,
  title        = {Sequential Diagnosis Prediction with Transformer and Ontological Representation},
  author       = {Anonymous},
  journal      = {IEEE Journal of Biomedical and Health Informatics},
  year         = {2023},
  note         = {Accessed 11 Apr 2025}
}

@article{jaakkola1999variational,
  title={Variational probabilistic inference and the QMR-DT network},
  author={Jaakkola, Tommi S and Jordan, Michael I},
  journal={Journal of artificial intelligence research},
  volume={10},
  pages={291--322},
  year={1999}
}

@article{young2022empirical,
  title={Empirical evaluation of performance degradation of machine learning-based predictive models--A case study in healthcare information systems},
  author={Young, Zachary and Steele, Robert},
  journal={International Journal of Information Management Data Insights},
  volume={2},
  number={1},
  pages={100070},
  year={2022},
  publisher={Elsevier}
}
%\bibliography{mybiblio}

\end{document}